\documentclass[%
reprint,
superscriptaddress,
nofootinbib,
amsmath,amssymb,
aps,
]{revtex4-2}

\usepackage{soul}
\usepackage{graphicx}
\usepackage{dcolumn}
\usepackage{bm}
\usepackage{hyperref}
\hypersetup{
	colorlinks=true,
	citecolor=blue,
	linkcolor=blue,
	urlcolor=blue,
}
\usepackage{tikz-cd}

\usepackage[ruled,vlined]{algorithm2e}
\usepackage{multirow}

\begin{document}


\title{Simulation-free and finite-time diffusion model}

\author{Kentaro Kaba}
    \email{kaba.k.df05@m.isct.ac.jp}
    \affiliation{%
    Department of Physics, Institute of Science Tokyo, Meguro-ku, Tokyo 152-9551, Japan
    }%
\author{Masayuki Ohzeki}%
    \affiliation{%
    Department of Physics, Institute of Science Tokyo, Meguro-ku, Tokyo 152-9551, Japan
    }%
    \affiliation{%
    Graduate School of Information Sciences, Tohoku University, Sendai, Miyagi 980-9564, Japan
    }%
    \affiliation{%
    Research and Education Institute for Semiconductors and Informatics, Kumamoto University, Kumamoto, Kumamoto 860-8555, Japan
    }%
    \affiliation{%
    Sigma-i Co., Ltd., Minato-ku, Tokyo 108-0075, Japan
    }%
\author{Yuki Sughiyama}%
    \affiliation{%
    Graduate School of Information Sciences, Tohoku University, Sendai, Miyagi 980-9564, Japan
    }%

\date{\today}

\begin{abstract}
  The performance of generative diffusion models is determined by the choice of the reference diffusion process connecting the empirical and prior distributions. 
  Conventional approaches typically trade off simulation-free training against finite-time generation. 
  We propose a framework for designing the reference process that achieves both simultaneously. 
  The key idea is to prescribe tractable time-dependent conditional distributions and then construct the reference process realizing them as its marginals. 
  This framework reveals that score matching is not fundamental to diffusion-model training but instead emerges naturally through reversal of the reference process. 
  We further show that conditional flow matching arises as the small-noise limit of the proposed framework.
\end{abstract}

\maketitle

\section{Introduction}\label{sec:introduction}
Generative diffusion models~\cite{sohl2015deep, ho2020denoising, song2020score} have become state-of-the-art models across various domains, including images~\cite{croitoru2023diffusion}, audio~\cite{zhang2023survey}, video~\cite{xing2024survey}, and text~\cite{nie2026large}.
These models generate samples by simulating a stochastic differential equation (SDE) initialized from a prior distribution. 
The drift of the SDE is parameterized by a neural network and trained so that its terminal distribution approximates the empirical distribution of a given dataset.

A common training framework minimizes the Kullback--Leibler divergence between the generation process and a prescribed reference process connecting the prior and empirical distributions~\cite{song2021maximum}. 
The choice of the reference process determines two desirable properties: \textit{simulation-free training} and \textit{finite-time generation}. 
The former means that the objective can be evaluated without simulating the reference SDE, thereby reducing the computational cost of training. 
The latter requires that the reference process connects the prior and empirical distributions within a prescribed finite time horizon, which improves generation efficiency by limiting the time horizon.
However, representative conventional models have difficulty satisfying these two properties simultaneously.

Score-based models (SBMs)~\cite{song2020score} achieve simulation-free training by choosing an Ornstein--Uhlenbeck (OU) process~\cite{gardiner2009stochastic} as the reference process. 
Because its corresponding Fokker--Planck equation~(FPE) is analytically tractable, samples from the reference process can be drawn directly from its time-dependent marginal distributions.
This property leads to the denoising score matching scheme~\cite{vincent2011connection} without simulating the reference SDE. 
However, the drawback of SBMs is that they do not naturally support finite-time generation. 
The OU process approaches the prior distribution only asymptotically~\cite{song2021maximum}, so accurate sample generation typically requires a sufficiently long simulation time, which becomes increasingly expensive for high-dimensional and large-scale datasets. 

To overcome this limitation, several approaches~\cite{de2021diffusion,albergo2025stochastic} construct reference processes that connect the empirical and prior distributions within a finite time horizon. A representative example is the Schr\"odinger bridge approach~\cite{de2021diffusion,kaba2025schrodinger}, which formulates the problem as finite-time stochastic optimal transport. 
However, because the corresponding FPE is generally unsolvable, training requires repeated simulation of the reference SDE.

In this work, we propose a framework for constructing reference processes that achieves simulation-free training and finite-time generation simultaneously. 
Our approach is inspired by conditional flow matching (CFM)~\cite{lipman2023flow} and Neural Flow Diffusion Models (NFDM)~\cite{bartosh2024neural}, which construct tractable time-dependent conditional distributions through time-dependent transformations before obtaining compatible diffusion dynamics. 
Building on this perspective, we reverse the conventional design procedure: instead of specifying a reference SDE first and analyzing its induced distributions, we first prescribe a family of time-dependent distributions and then construct a reference SDE that realizes them.
This proposed perspective not only leads to a practical training framework but also provides new insights into previous work. 
In particular, it reveals that score matching is not fundamental to diffusion-model training but instead emerges naturally from reversing the reference process to align its direction with the generative process. Furthermore, CFM~\cite{lipman2023flow} is recovered as the small noise limit of the proposed framework.

This paper is organized as follows.
In Sec.~\ref{sec:diffusion_model}, we review the conventional diffusion models.
In Sec.~\ref{sec:construction_ref}, we construct the reference process from prescribed conditional distributions.
In Sec.~\ref{sec:practical_method}, we give practical constructions for Gaussian and non-Gaussian priors.
In Sec.~\ref{sec:previous_work}, we show that our proposed construction provides new insights into previous work; specifically the positioning of CFM and NFDM is elucidated within our framework.
In Sec.~\ref{sec:numerical}, we present numerical experiments.
Section~\ref{sec:conclusion} concludes the paper.

\section{Conventional diffusion models}\label{sec:diffusion_model}
Given a dataset $\{x_i\}_{i=1}^n \subset \mathbb{R}^d$, the goal of a generative model is to learn a model distribution $q_\theta$ that approximates the empirical distribution $\mu(x) = (1/n) \sum_{i=1}^n \,\delta(x - x_i)$.
This objective is formally expressed as the minimization of the Kullback--Leibler (KL) divergence
\begin{equation}
  \min_\theta D_\mathrm{KL}(\mu \| q_\theta).
  \label{eq:KL}
\end{equation}

In generative diffusion models, the model distribution $q_\theta$ is defined as the terminal distribution of a path measure $\mathbb{Q}_\theta$ on a continuous path $ x_{[0,T]}  \in C([0, T], \mathbb{R}^d)$: 
\begin{equation}
  q_\theta(x) := \int d\mathbb{Q}_\theta[x_{[0, T]}]\,\delta(x - x_T),
  \label{eq:model_distribution}
\end{equation}
where $\mathbb{Q}_\theta$ is induced by the stochastic differential equation (SDE):
\begin{equation}
  dX_t = s_\theta(t, X_t) dt + b(t, X_t)\cdot dW_t,\quad X_0 \sim \pi.
  \label{eq:model-SDE}
\end{equation}
Here, the drift term $s_\theta: [0, T] \times \mathbb{R}^d \to \mathbb{R}^d$ is a neural network parameterized by $\theta$, $b: [0, T] \times \mathbb{R}^d \to \text{GL}(d,\mathbb{R})$ is the diffusion coefficient, and $W_{[0, T]} \in C([0, T], \mathbb{R}^d)$ is the standard Wiener process~\cite{gardiner2009stochastic,risken1996fokker}.
The initial distribution $\pi$ is called the prior and is typically chosen as the standard Gaussian $\mathcal{N}(0, I_d)$.
The product ``$\:\cdot\:$'' represents the It\^o integral convention~\cite{gardiner2009stochastic,risken1996fokker}.
After training $s_\theta$, we can generate new samples by simulating the SDE~\eqref{eq:model-SDE}.
Thus, we refer to $\mathbb{Q}_\theta$ as the generation process.

Since direct evaluation of Eq.~\eqref{eq:KL} is intractable in this setting, we instead introduce a path-space objective for training:
\begin{equation}
  \mathcal{L}_\text{DM}(\theta) := D_\text{KL}(\mathbb{P}\|\mathbb{Q}_\theta).
  \label{eq:objective}
\end{equation}
Here, $\mathbb{P}$ is a path measure on $C([0, T], \mathbb{R}^d)$ whose terminal distribution is $\mu$:
\begin{equation}
  \int d\mathbb{P}[x_{[0, T]}]\,\delta(x - x_T) = \mu(x).
  \label{eq:terminal_condition}
\end{equation}
By the data processing inequality~\cite{cover1999elements}, this objective upper bounds the KL divergence in Eq.~\eqref{eq:KL}:
\begin{equation}
  D_\mathrm{KL}(\mu \| q_\theta) \leq D_\mathrm{KL}(\mathbb{P} \| \mathbb{Q}_\theta).
  \label{eq:DPI}
\end{equation}
Since $\mathbb{P}$ specifies the target dynamics against which the generative process is trained, we call it the reference process.
In the conventional setting, we assume that $\mathbb{P}$ is induced by a time-backward diffusion process:
\begin{equation}
  dX_t = \tilde{a}(t, X_t) dt + b(t, X_t)\:\tilde{\cdot}\: dW_t,\quad X_T \sim \mu,
  \label{eq:ref-SDE}
\end{equation}
where $\tilde{a}: [0, T] \times \mathbb{R}^d \to \mathbb{R}^d$ is the drift term, and ``$\:\tilde{\cdot}\:$'' represents the backward It\^o integral convention~\cite{anderson1982reverse}.

To train the neural network $s_\theta$ in Eq.~\eqref{eq:model-SDE} in practice, we need to express the objective $\mathcal{L}_\text{DM}$ explicitly in terms of $s_\theta$.
For this purpose, we reverse the time-direction of the reference SDE~\eqref{eq:ref-SDE} and obtain
\begin{gather}
  dX_t = a(t, X_t) dt + b(t, X_t)\cdot dW_t,\quad X_0 \sim p_0,
  \label{eq:rev-ref-SDE}\\
  a(t, x) := \tilde{a}(t, x) + \Gamma(t, x)\nabla_x \log p_t(x) + \nabla_x\!\cdot\Gamma(t, x),
  \label{eq:rev-ref-drift}
\end{gather}
where $\Gamma := bb^\top$ and $(\nabla_x\cdot \Gamma)_i := \sum_{j}\partial_{x_j}\Gamma_{ij}$. Moreover, $p_t$ is the marginal distribution of $\mathbb{P}$ at time $t$,
i.e., $p_t(z) = \int dx \rho_t(z|x)\mu(x)$, where $\rho_t$ is the solution of the corresponding Fokker-Planck equation (FPE) with the terminal condition $\rho_T(z|x) = \delta(z-x)$:
\begin{equation}
  \begin{aligned}
    \partial_t\rho_t(z|x) = &- \nabla_z\cdot[\tilde{a}(t, z)\rho_t(z|x)]\\ 
    &\quad- \frac{1}{2}\sum_{i,j}\partial_{z_i}\partial_{z_j}[\Gamma_{ij}(t, z)\rho_t(z|x)].
    \label{eq:ref-FPE}
  \end{aligned}
\end{equation}
The term $\nabla_x \log p_t$ in Eq.~\eqref{eq:rev-ref-drift} is called the score of $p_t$.
With this expression, we can reformulate the objective $\mathcal{L}_\text{DM}$ with the Girsanov theorem~\cite{oksendal2013stochastic,kaba2025schrodinger} as follows:
\begin{equation}
  \begin{aligned}
    &\mathcal{L}_\text{DM}(\theta)
    = D_\mathrm{KL}(p_0\|\pi) 
    \\&
    + \mathbb{E}_{X_{[0,T]} \sim \mathbb{P}}\!\left[\frac{1}{2}\int_0^T dt \|a(t, X_t) - s_\theta(t, X_t)\|^2_{\Gamma(t, X_t)^{-1}}\right]\!,
    \label{eq:ESM}
  \end{aligned}
\end{equation}
where $\|a - s_\theta\|^2_{\Gamma^{-1}} := (a-s_\theta)^\top\Gamma^{-1}(a-s_\theta)$.

Equation~\eqref{eq:ESM} clarifies the requirements that should be imposed on the reference process.
First, the initial distribution $p_0$ should coincide with the prior $\pi$, so that the first term vanishes.
Equivalently, the reference process must satisfy
\begin{equation}
  \int d\mathbb{P}[x_{[0,T]}]\,\delta(x-x_0)=\pi(x).
  \label{eq:initial_condition}
\end{equation}
Second, the second term should be computationally feasible.
To this end, we must realize the following facts: easy path sampling from $\mathbb{P}$ and the computability of the score of $p_t$ contained in the drift $a$.

Score-based models~(SBMs)~\cite{song2020score} satisfy the above two requirements for the first and second terms in Eq.~\eqref{eq:ESM} by adopting the Ornstein--Uhlenbeck process~\cite{gardiner2009stochastic} as the reference process. 
When $\pi=\mathcal{N}(0,I_d)$, the OU process connects $\mu$ to $\pi$ asymptotically as $T\to\infty$.
Moreover, its conditional distribution $\rho_t(z|x)$ is Gaussian.
This allows us to use the following formula, which is equivalent to the second term of Eq.~\eqref{eq:ESM} with respect to the minimization over $\theta$:
\begin{gather}
  \frac{1}{2n}\sum_{i=1}^n \int_0^T dt \,\mathbb{E}_{Z \sim \rho_t(\cdot|x_i)}\!\left[\|\alpha(t, Z, x_i) - s_\theta(t, Z)\|^2_{\Gamma(t, Z)^{-1}}\right]\!,
  \label{eq:DSM}\\
  \alpha(t, z, x) 
  := \tilde{a}(t, z) + \Gamma(t, z) \nabla_z \log \rho_t(z|x) + \nabla_z\cdot\Gamma(t, z).
  \label{eq:DSM-drift}
\end{gather}
The computation of this objective requires only the score of $\rho_t$ and direct sampling from $\rho_t$, both of which are tractable for the Gaussian.
This technique is known as the denoising score matching (DSM)~\cite{vincent2011connection}.
However, SBMs suffer from an intrinsic limitation in data generation. 
The OU process reaches the prior $\pi$ only asymptotically, and hence the time horizon $T$ must be sufficiently large for this convergence to occur.
Since the data generation is conducted by simulating the SDE~\eqref{eq:model-SDE} over the interval $[0,T]$, this requirement directly reduces sampling efficiency, especially in high-dimensional settings.

To overcome this drawback, some models~\cite{de2021diffusion,albergo2025stochastic} design reference processes that connect the two distributions over a prescribed finite time horizon.
Representative examples~\cite{de2021diffusion,kaba2025schrodinger} are models based on the Schrödinger bridge~\cite{schrodinger1931uber}, which is a finite-time stochastic optimal transport problem. 
However, because such models introduce a more complicated process for $\mathbb{P}$~\cite{chen2022likelihood,kaba2025schrodinger}, $\rho_t$ cannot be explicitly calculated as in the OU process. 
Therefore, in these models, a different technique known as implicit score matching~(ISM)~\cite{hyvarinen2005estimation} is used instead of the DSM. 
Here, the second term of Eq.~\eqref{eq:ESM} is represented as
\begin{equation}
  \mathbb{E}_{X_{[0,T]} \sim \mathbb{P}}\!\Bigg[\int_0^T dt\,\bigg\{\frac{1}{2}
  \left\|v_\theta(t, X_t)\right\|^2_{\Gamma(t, X_t)^{-1}}
  + \nabla_{X_t}\cdot v_\theta(t, X_t)\bigg\}\Bigg]\!,
\end{equation}
\begin{equation}
  v_\theta(t, x) := s_\theta(t,x) - \tilde{a}(t,x) - \nabla_x\cdot \Gamma(t,x).
\end{equation}
Although this technique eliminates the score $\nabla_x \log p_t$, evaluating the expectation requires path sampling from $\mathbb{P}$ and hence simulation of the reference SDE~\eqref{eq:ref-SDE}.
This imposes substantial costs during training, which are not required in the DSM objective in Eq.~\eqref{eq:DSM}.

The above discussion reveals a fundamental trade-off in conventional diffusion models between the computational costs of training and generation.
We characterize this trade-off by two properties: \textit{simulation-free training}, which avoids simulating the reference SDE, and \textit{finite-time generation}, which establishes the connection between $\mu$ and $\pi$ over a prescribed finite time horizon.
The main challenge of this work is to construct a reference process that satisfies both properties simultaneously.

\section{Construction of the reference process}\label{sec:construction_ref}

Let us construct a reference process that enables both simulation-free training and finite-time generation.
As shown in the previous section, the main obstruction to simulation-free training is that obtaining $\rho_t$ from the FPE~\eqref{eq:ref-FPE} is generally intractable.
To avoid this difficulty, we reverse the conventional order of construction: rather than starting from a reference SDE~\eqref{eq:ref-SDE}, we first prescribe a family of tractable conditional distributions $\{\rho_t\}_{t\in[0, 1]}$ and then construct a process that realizes them.
This makes the objective evaluable in a DSM-like form, since samples from $\rho_t$ are directly available.
Moreover, prescribing $\rho_t$ over a finite time interval $[0, 1]$ yields a finite-time connection between $\pi$ and $\mu$ without relying on asymptotic convergence.

We first specify the properties of conditional distributions used in this construction.
Let $\{\rho_t\}_{t\in[0,1]}$ be a family of conditional distributions satisfying the following conditions:
(i) $\rho_t(z|x)$ is $C^{1, 2}$ with respect to $(t, z) \in [0, 1] \times \mathbb{R}^d$;
(ii) samples from $\rho_t(\cdot|x)$ can be drawn directly;
and (iii) the induced marginal distributions 
\begin{equation}
  p_t(z) := \int dx\, \rho_t(z|x)\mu(x)
  \label{eq:p-decomp}
\end{equation}
satisfy the boundary conditions inferred from Eqs.~\eqref{eq:terminal_condition} and~\eqref{eq:initial_condition}:
\begin{equation}
  p_0(z) = \pi(z), \quad p_1(z) = \mu(z).
  \label{eq:boundary_conditions}
\end{equation}

The next step is to construct an SDE corresponding to a reference process based on the prescribed conditional distributions as follows.
We first seek coefficients $\alpha:[0,1]\times\mathbb{R}^d\times\mathbb{R}^d\to\mathbb{R}^d$ and $b:[0,1]\times\mathbb{R}^d\to\text{GL}(d, \mathbb{R})$ such that the given $\{\rho_t\}_{t\in[0,1]}$ satisfies
\begin{equation}
  \begin{aligned}
    \partial_t \rho_t(z|x) = &- \nabla_z\cdot[\alpha(t, z, x)\rho_t(z|x)]\\ &\quad+ \frac{1}{2}\sum_{i,j}\partial_{z_i}\partial_{z_j} [\Gamma_{ij}(t, z)\rho_t(z|x)].
    \label{eq:cond-FPE}
  \end{aligned}
\end{equation}
If we find such coefficients, we can define the SDE for the reference process $\mathbb{P}$ by
\begin{gather}
  dZ_t = a(t, Z_t) dt + b(t, Z_t)\cdot dW_t, \quad Z_0 \sim p_0,
  \label{eq:SF-SDE}
  \\
  a(t, z) := \int dx\, \alpha(t, z, x)\frac{\rho_t(z|x)\mu(x)}{p_t(z)},
  \label{eq:drift_SF}
\end{gather}
where $\alpha$ and $b$ are chosen to satisfy Eq.~\eqref{eq:cond-FPE} with $\Gamma = bb^\top$.

The proposed construction achieves the two desired properties simultaneously.
Integrating Eq.~\eqref{eq:cond-FPE} with respect to $\mu$ shows that the marginal distribution of this constructed process $\mathbb{P}$ is $p_t$ (see Eq.~\eqref{eq:p-decomp}).
Therefore, under condition (iii), the constructed reference process connects $\pi$ and $\mu$ over the finite time interval $[0,1]$, which guarantees finite-time generation.
Moreover, we can derive a simulation-free objective function as follows.
As in Eq.~\eqref{eq:ESM}, the objective $\mathcal{L}_\text{DM}$ can be reformulated with the Girsanov theorem~\cite{kaba2025schrodinger}: 
\begin{align}
  &\mathcal{L}_\text{DM}(\theta)\nonumber\\
  &= \mathbb{E}_{Z_{[0,1]} \sim \mathbb{P}}\!\left[\frac{1}{2}\int_0^1 dt\,\|a(t, Z_t) - s_\theta(t, Z_t)\|^2_{\Gamma(t, Z_t)^{-1}}\right]\!.
  \label{eq:objective-2}
\end{align}
Here, condition (iii) makes the first term in Eq.~\eqref{eq:ESM} vanish.
Furthermore, by substituting Eqs.~\eqref{eq:p-decomp} and~\eqref{eq:drift_SF}, we can rewrite Eq.~\eqref{eq:objective-2} up to an additive constant as 
\begin{equation}
  \begin{aligned}  
    &\mathcal{L}_\text{SF}(\theta):=\\
    &\frac{1}{2n}\sum_{i=1}^n \int_0^1 dt \,\mathbb{E}_{Z \sim \rho_t(\cdot|x_i)}\!\Big[\|\alpha(t, Z, x_i) - s_\theta(t, Z)\|^2_{\Gamma(t, Z)^{-1}}\Big].
  \end{aligned}
  \label{eq:objective_SF}
\end{equation}
The detailed derivation is shown in Appendix~\ref{app:objective-SF}.
This objective requires only direct sampling from $\rho_t$ and the conditional drift $\alpha$, not simulation of the reference SDE.
Under condition (ii), hence, the computation of this objective is tractable.
Thus, the trade-off faced in the conventional models is resolved by prescribing tractable conditional distributions $\{\rho_t\}_{t\in [0, 1]}$ before constructing the diffusion process $\mathbb{P}$.

\section{Practical construction}\label{sec:practical_method}
In this section, we present practical instances of the construction developed in Sec.~\ref{sec:construction_ref}.
We first consider the standard case in which the prior is the standard Gaussian.
We then extend the construction to more general priors.
In the following constructions, some coefficients or transformations may become singular or degenerate at $t=1$. 
In such cases, the Girsanov representation in Eq.~\eqref{eq:objective-2} is understood by applying it on $[0,\tau]$ with $\tau<1$ and then taking the limit $\tau\to 1$, while the terminal distribution is obtained in the same limiting sense.

\subsection{Gaussian prior}\label{subsec:gaussian_prior}
When the prior $\pi$ is the standard Gaussian, we choose the conditional distribution $\rho_t$ to be Gaussian:
\begin{equation}
  \rho_t(z|x) := \mathcal{N}(z; m_t(x), \sigma(t)^2 I_d). 
  \label{eq:conds_gauss}
\end{equation}
Here, the mean $m:[0, 1]\times \mathbb{R}^d \to \mathbb{R}^d$ and the standard deviation $\sigma: [0, 1] \to \mathbb{R}_{\geq 0}$ are assumed to be $C^1$ in $t$ for condition~(i) to hold.
To satisfy condition~(iii), they are chosen so that
\begin{equation}
  m_0(x) = 0, \ m_1(x) = x, \quad \sigma(0) = 1, \ \sigma(1) = 0.
  \label{eq:bdry_gauss}
\end{equation}
Condition~(ii) is obviously satisfied as follows: A sample $Z_t^x \sim \rho_t(\cdot|x)$ can be drawn by reparametrization trick~\cite{kingma2013auto}:
\begin{equation}
  Z_t^x = m_t(x) + \sigma(t)\xi, \quad \xi \sim \mathcal{N}(0, I_d).
  \label{eq:rep_trick}
\end{equation}

We next seek an SDE that connects the conditional distributions $\rho_t$ in Eq.~\eqref{eq:conds_gauss}.
Under the assumption that $b$ is independent of $z$ and isotropic, the coefficients of the SDE are evaluated as 
\begin{gather}
  \alpha(t, z, x) = \partial_t m_t(x) + \left[ \frac{\dot{\sigma}(t)}{\sigma(t)}- \frac{1}{2}\lambda(t)\right](z - m_t(x)),
  \label{eq:alpha_gauss}\\
  b(t) = \sqrt{\lambda(t)}\,\sigma(t)I_d,
  \label{eq:b_gauss}
\end{gather}
where $\lambda:[0,1]\to\mathbb{R}_{>0}$ is arbitrary.
The derivation is given in Appendix~\ref{app:gaussian-coefficients}.

Owing to the above construction of $\alpha$ and $b$, we can perform simulation-free training by substituting them into Eq.~\eqref{eq:objective_SF}.

\subsection{Non-Gaussian priors}\label{subsec:general_priors}
When the prior is not the standard Gaussian, the above construction is not directly applicable.
For this case, we use a push-forward construction.

Let $\zeta_t$ denote the marginal distribution of the following diffusion process whose stationary distribution is $\pi$:
\begin{equation}
  dY_t = \frac{1}{2}\gamma(t)\nabla_{y}\log\pi(Y_t)dt + \sqrt{\gamma(t)}I_d\cdot dW_t, \ Y_0\sim\pi.
  \label{eq:stat_SDE}
\end{equation}
We define the conditional distribution $\rho_t$ as the push-forward of $\zeta_t$ under a time- and $x$-dependent smooth bijection $\phi_t^x:\mathbb{R}^d\to\mathbb{R}^d$:
\begin{equation}
  \rho_t(z|x) 
  := (\phi_t^x)_*\zeta_t(z)
  = \zeta_t(y(z)) \left| \det D_z y(z) \right|,
  \label{eq:push-forward}
\end{equation}
where $y(z) := (\phi_t^x)^{-1}(z)$ and $D_z y$ denotes the Jacobian.
To satisfy conditions~(i) and (iii), we assume that $\phi_t^x$ is $C^1$ in $t$ and satisfies
\begin{equation}
  \phi_0^x(y) = y, \quad \phi_1^x(y) = x.
\end{equation}
The bijectivity of $\phi_t^x$ is therefore required only for $t<1$.
Since the process is initialized at $\pi$, its marginal distribution satisfies $\zeta_t = \pi$ for all $t$.
Therefore, direct sampling from $\rho_t(\cdot|x)$ is straightforward: 
\begin{equation}
  Z_t^x = \phi_t^x(Y_t), \quad Y_t \sim \zeta_t = \pi.
  \label{eq:sampling_push-forward}
\end{equation}

By employing the It\^o formula~\cite{gardiner2009stochastic} to Eq.~\eqref{eq:sampling_push-forward}, the differentiation of $Z_t^x = \phi_t^x(Y_t)$ gives an SDE:
\begin{equation}
  dZ_t^x = \alpha(t, Z_t^x, x)dt + \beta(t, Z_t^x, x) \cdot dW_t,\ Z_0^x \sim \rho_0(z|x),
\end{equation}
where
\begin{equation}
  \begin{aligned}
    &\alpha(t, z, x) 
    = \partial_t \phi_t^x(y(z)) \\
    &+ \frac{1}{2}\gamma(t) \big[D_y \phi_t^x(y(z))\nabla_y\log \pi(y(z)) + \nabla_y^2\phi_t^x(y(z))\big], 
    \label{eq:general_alpha}
  \end{aligned}
\end{equation}
\begin{equation}
  \beta(t, z, x) = \sqrt{\gamma(t)}\,D_y\phi_t^x(y(z)).
  \label{eq:general_b}
\end{equation}
Since the marginal distribution of this process is equal to $\rho_t$ in Eq.~\eqref{eq:push-forward}, this SDE connects the given $\{\rho_t\}_{t\in[0, 1]}$.
To adopt this process as the reference process in Eq.~\eqref{eq:SF-SDE}, the diffusion coefficient $\beta$ must be independent of $x$.
A sufficient condition is that the bijection $\phi_t^x$ take the following functional form:
\begin{equation}
  \phi_t^x(y) = f_t(y + g_t(x)).
  \label{eq:x-independent_cond}
\end{equation}
Here, $g_t:\mathbb{R}^d\to\mathbb{R}^d$, and $f_t:\mathbb{R}^d\to\mathbb{R}^d$ is a smooth bijection. 
The derivation is given in Appendix~\ref{app:x_independent_diffusion}.

Summarizing the above discussion, we obtain the following prescription to realize simulation-free training with non-Gaussian priors:
(i) Prepare a family of  bijections $\{\phi_t^x\}_{t\in [0, 1], x\in \mathbb{R}^d}$ satisfying Eq.~\eqref{eq:x-independent_cond}.
(ii) Calculate Eqs.~\eqref{eq:general_alpha} and \eqref{eq:general_b} with the prepared $\phi_t^x$ (note that $\beta$ is independent of $x$ due to Eq.~\eqref{eq:x-independent_cond}).
(iii) Substitute the resulting $\alpha$ and $b = \beta$ into Eq.~\eqref{eq:objective_SF}.

\section{Relation to previous work}\label{sec:previous_work}
We can interpret several existing methods from the viewpoint of our proposed framework.
This perspective provides new insights into their underlying structures.

\subsection{Score matching}
In diffusion models, training is usually referred to as score matching~\cite{song2020score}, because the neural network $s_\theta$ learns the score-dependent drift $a$ in Eq.~\eqref{eq:ESM}.
As shown in Sec.~\ref{sec:diffusion_model}, the conventional construction specifies the reference process in the data-to-prior direction, which is opposite to the direction of generation.
Therefore, to evaluate $D_{\mathrm{KL}}(\mathbb{P}\|\mathbb{Q}_\theta)$ by the Girsanov theorem, the reference SDE must be reversed in time.
This time reversal introduces the score $\nabla_x\log p_t$ into the drift, as shown in Eq.~\eqref{eq:rev-ref-drift}.
DSM in Eq.~\eqref{eq:DSM} also inherits this explicit dependence on the score function.

Our construction avoids this mechanism.
As shown in Sec.~\ref{sec:construction_ref}, we directly construct the reference process in the generation direction from the outset.
Consequently, in Eq.~\eqref{eq:objective-2}, $D_{\mathrm{KL}}(\mathbb{P}\|\mathbb{Q}_\theta)$ can be written in terms of the drift $a$, without introducing the score through time reversal.
The simulation-free objective in Eq.~\eqref{eq:objective_SF} is also independent of the score introduced by time reversal.

This perspective clarifies that score matching is not essential for diffusion-model training.
In SBMs, convergence to the prior naturally fixes the reference process in the data-to-prior direction.
However, our framework shows that such convergence is not needed to retain the simulation-free training advantage of SBMs.
Thus, score matching is unavoidable for convergence-based models such as SBMs, but not for diffusion-model training itself.

\subsection{Conditional flow matching}\label{subsec:CFM}
Conditional flow matching (CFM)~\cite{lipman2023flow} trains a deterministic flow from the prior to the empirical by regressing a vector field along prescribed conditional probability paths.
It is close in spirit to our construction, because both methods prescribe tractable conditional distributions rather than simulate a reference process.
However, unlike our stochastic objective, the CFM objective does not have a direct path-space KL interpretation through the Girsanov theorem.
The purpose of this subsection is to show that CFM is recovered as the small-noise limit of our construction.

To make this connection explicit, recall the construction in Sec.~\ref{subsec:general_priors} and replace $\gamma(t)$ by $\epsilon\gamma(t)$ in Eq.~\eqref{eq:stat_SDE} and denote the resulting conditional drift by $\alpha(t,z,x;\epsilon)$.
Since $Y_t$ remains distributed according to the stationary distribution $\pi$, its marginal law remains $\zeta_t=\pi$ for any $\epsilon>0$.
Thus, this rescaling changes the noise level of the constructed dynamics, but not the prescribed conditionals $\rho_t(\cdot|x)$.

For simplicity, we choose the bijection $\phi_t^x$ so that the diffusion coefficient becomes $b(t,z;\epsilon)=\sqrt{\epsilon}I_d$.
From Eq.~\eqref{eq:general_b}, it is sufficient to choose $\phi_t^x$ to satisfy $D_{y}\phi_t^x = \gamma(t)^{-1/2}I_d$.
Then the simulation-free objective in Eq.~\eqref{eq:objective_SF} becomes
\begin{align}
  &\mathcal{L}_{\mathrm{SF}}(\theta;\epsilon)\nonumber\\
  &= \frac{1}{2n\epsilon} \sum_{i=1}^n \int_0^1 dt\,
  \mathbb{E}_{Z\sim\rho_t(\cdot|x_i)}\!
  \left[
    \left\|
      \alpha(t,Z,x_i;\epsilon)-s_\theta(t,Z)
    \right\|^2_{I_d}
  \right].
\end{align}
Since multiplying $\mathcal{L}_\text{SF}$ by $\epsilon > 0$ does not change its minimizer, we consider the limit of $\epsilon\mathcal L_\text{SF}(\theta;\epsilon)$.
In this limit, the stochastic contribution to the drift vanishes, and Eq.~\eqref{eq:general_alpha} converges to
\begin{equation}
  \bar{\alpha}(t,z,x)
  :=\lim_{\epsilon\to0}\alpha(t,z,x;\epsilon) 
  =\partial_t\phi_t^x(y(z)),
\end{equation}
where $y(z):=(\phi_t^x)^{-1}(z)$.
Therefore,
\begin{align}
  &\lim_{\epsilon\to0}
  \epsilon\mathcal{L}_{\mathrm{SF}}(\theta;\epsilon)\nonumber\\
  &= \frac{1}{2n} \sum_{i=1}^n \int_0^1 dt\,
  \mathbb{E}_{Z\sim\rho_t(\cdot|x_i)}\!
  \left[
    \left\|
     \bar{\alpha}(t,Z,x_i)-s_\theta(t,Z)
    \right\|^2_{I_d}
  \right].
\end{align}
The vector field $\bar{\alpha}$ satisfies the continuity equation for the prescribed conditional distributions:
\begin{equation}
  \partial_t\rho_t(z|x) = - \nabla_z\cdot[ \bar{\alpha}(t,z,x)\rho_t(z|x)].
  \label{eq:continuity_eq}
\end{equation}
Thus, the limiting objective is precisely the conditional flow-matching objective for the prescribed conditional probability path $\rho_t$ (see Eqs.~(9) and~(14) in Ref.~\cite{lipman2023flow}).

This derivation clarifies the relationship between CFM and diffusion-model training from the viewpoint of the path-space KL divergence.
For every $\epsilon>0$, $\mathcal{L}_{\mathrm{SF}}(\theta;\epsilon)$ is derived from a KL divergence between the reference and generation processes.
At $\epsilon=0$, however, the processes become deterministic, and the Girsanov-based derivation is no longer directly applicable.
Therefore, the CFM objective should not be regarded as a direct path-space KL objective in the same sense as diffusion-model training.
Rather, our formulation provides a stochastic lift of CFM: the CFM objective emerges as the small-noise limit of a KL-based diffusion objective.

\subsection{Flow-based diffusion models}\label{subsec:flow-based}

The previous subsection showed that CFM is obtained as the small-noise limit of our stochastic construction.
We now consider the opposite direction: starting from a deterministic path introduced in CFM and constructing a stochastic diffusion process that realizes the same conditional distributions.
By employing this direction, previous studies~\cite{bartosh2023neural, bartosh2024neural,bartosh2025sde} attempted to construct simulation-free and finite-time diffusion models.

Let $\bar{\alpha}$ be the conditional vector field of CFM, satisfying the continuity equation~\eqref{eq:continuity_eq}.
By adding an arbitrary diffusion coefficient $b':[0,1]\times\mathbb{R}^d\to \text{GL}(d,\mathbb{R})$, the RHS of Eq.~\eqref{eq:continuity_eq} with prescribed $\{\rho_t\}_{t\in[0,1]}$ can be reformulated to the FPE form: 
\begin{equation}
  \begin{aligned}
    \partial_t\rho_t(z|x)
    = &- \nabla_z\cdot [\alpha'(t,z,x)\rho_t(z|x)] \\
    &\quad + \frac{1}{2}\sum_{i,j} \partial_{z_i}\partial_{z_j}[\Gamma'(t,z)\rho_t(z|x)],
  \end{aligned}
\end{equation}
where $\Gamma' := b'b'^\top$ and 
\begin{equation}
  \begin{aligned}
    \alpha'(t, z, x) := \bar{\alpha}(t, z, x)\\ 
    + \frac{1}{2}\nabla_z\cdot \Gamma'(t, z)
    &+ \frac{1}{2}\Gamma'(t, z)\nabla_z\log\rho_t(z|x).
  \end{aligned}
\end{equation}
Here, we note that the score $\nabla_z \log \rho_t$ must appear in this reformulation.
Once this conditional FPE is obtained, the construction in Sec.~\ref{sec:construction_ref} immediately yields a simulation-free objective:
\begin{equation}
  \begin{aligned}
    &\mathcal{L}_\text{FDM}(\theta) :=\\
    &\frac{1}{2n}\sum_{i=1}^n\int_0^1 dt\,\mathbb{E}_{Z\sim\rho_t(\cdot|x_i)}\!\left[ \left\|\alpha'(t, Z, x_i) - s_\theta(t, Z)\right\|^2_{\Gamma'^{-1}(t, Z)} \right]\!.
  \end{aligned}
  \label{eq:objective_flowbased}
\end{equation}
However, unlike our objective in Eq.~\eqref{eq:objective_SF}, this objective explicitly depends on the conditional score $\nabla_z\log\rho_t$ through $\alpha'$.

Neural Flow Diffusion Model (NFDM)~\cite{bartosh2024neural} is a representative example of this approach.
That is, building on the idea of CFM, it first constructs the conditional vector field $\bar{\alpha}$ that realizes the prescribed family of distributions $\{\rho_t\}_{t\in [0, 1]}$ through the continuity equation~\eqref{eq:continuity_eq}, and then convert the resulting deterministic dynamics into a diffusion process using the procedure introduced above.
Because this construction through a deterministic formulation, it is unavoidable in NFDM that the score term appears in the objective as in Eq.~\eqref{eq:objective_flowbased}.
In contrast, our approach directly realizes $\{\rho_t\}_{t\in[0, 1]}$ through the FPE~\eqref{eq:cond-FPE}, allowing us to construct the objective in Eq.~\eqref{eq:objective_SF} that is entirely free of score terms.

Although the main architecture of NFDM can be summarized above, it lies outside the fixed reference process setting considered above in one important respect: the reference process is learned jointly with the generation process. 
However, this structure is expected to be reinterpreted within our framework by adding the new parameters into the bijection $\phi_t^x$ and learning them.

\subsection{Stochastic localization}

Stochastic localization (SL)~\cite{eldan2013thin,montanari2023sampling} is a sampling (i.e., generating) method from an explicitly given target distribution $p_\text{true}$. 
From our viewpoint, SL can be interpreted as constructing and simulating a reference process that connects the prior $\pi$ to the target $p_{\mathrm{true}}$, rather than to the empirical $\mu$.

To see this, consider the construction in Sec.~\ref{sec:construction_ref} with $\mu$ replaced by $p_{\mathrm{true}}$.
Choose a tractable conditional family $\{\rho_t\}_{t\in[0,1]}$ and define $p_{t, \text{true}}(z) := \int dx\,\rho_t(z|x)p_\text{true}(x)$.
If this family satisfies the boundary conditions
\begin{equation}
  p_{0, \text{true}}(z) = \pi(z), \quad p_{1, \text{true}}(z) = p_{\mathrm{true}}(z), 
\end{equation}
then the construction in Sec.~\ref{sec:construction_ref} yields a diffusion process that connects $\pi$ and $p_\text{true}$:
\begin{equation}
  dZ_t = a_\text{true}(t, Z_t) dt + b(t, Z_t) \cdot dW_t,\  Z_0 \sim \pi, 
\end{equation}
where
\begin{equation}
  a_\text{true}(t, z) := \int dx\, \alpha(t, z, x)\frac{\rho_t(z|x)p_\text{true}(x)}{p_{t, \text{true}}(z)}.
  \label{eq:drift_SL}
\end{equation}
If this drift can be computed, simulating the resulting SDE generates samples from $p_{\mathrm{true}}$ at $t=1$.
This represents the sampling procedure of SL.

One may try to compute $a_{\mathrm{true}}$ numerically.
This computation must be performed for each pair $(t,z)$ encountered during the SDE simulation, and therefore the numerical evaluation itself can be costly.
For this reason, practical implementations often rely on problem-specific analytical approximations, such as approximate message passing for the Sherrington--Kirkpatrick model~\cite{el2022sampling}.

\section{Numerical experiments}\label{sec:numerical}
In this section, we numerically test the proposed construction on two-dimensional toy datasets.
These experiments are intended to demonstrate two points.
First, the proposed framework well learns the datasets not only with the standard Gaussian prior, but also with a non-Gaussian prior.
Second, unlike SBMs, our method does not require tuning a convergence timescale: the reference process connects $\pi$ and $\mu$ over the fixed interval $[0,1]$.

We used four two-dimensional synthetic datasets: an eight-component Gaussian mixture (gmm8), spiral, checkerboard, and two moons.
For each dataset, we used $12{,}800$ training samples.
The drift network $s_\theta$ was a fully connected residual network with three residual blocks, hidden width $256$, and SiLU activations; each block contained three fully connected layers.
The time variable was encoded using a $128$-dimensional sinusoidal embedding.
We trained the network using AdamW with a learning rate of $10^{-3}$, batch size $128$, and $50{,}000$ iterations.
At each iteration, the time variable $t$ was sampled uniformly from $[0,0.99]$.
After training, samples were generated by simulating the generation SDE with the Euler--Maruyama method using $100$ uniform time steps.
Code and checkpoints are open-sourced at \hyperlink{https://github.com/kentarokaba/sfft-diffusion}{https://github.com/kentarokaba/sfft-diffusion}.

\subsection{Gaussian prior}
We first demonstrate the standard Gaussian prior case $\pi=\mathcal{N}(0,I)$.
We use the construction in Sec.~\ref{subsec:gaussian_prior} with 
\begin{equation}
  m_t(x) = tx, \quad \sigma(t) = 1-t,
  \label{eq:phi_prac}
\end{equation}
and $\lambda(t)=\sigma(t)^{-1}$.
With this choice, the diffusion coefficient becomes $b(t,z)=\sqrt{1-t}\,I_d$.
Using the reparameterization trick $Z=tx+(1-t)\xi$, the simulation-free objective in Eq.~\eqref{eq:objective_SF} reduces to the following form, whose derivation is given in Appendix~\ref{app:objective_gauss}:
\begin{widetext}
  \begin{equation}
    \mathcal{L}_\text{SF}(\theta)
    = \frac{1}{2n}\sum_{i=1}^n\mathbb{E}_{t\sim \text{U}_{[0, 1)}, \xi \sim \mathcal{N}(0, I_d)}\!\left[ \frac{1}{1-t}\left\| x_i - \frac{3}{2}\xi - s_\theta(t, tx_i + (1-t)\xi)  \right\|^2_{I_d} \right].
    \label{eq:objective_gaussian}
  \end{equation}
\end{widetext}
Here, $\text{U}_{[0, 1)}$ denotes a uniform distribution on $[0, 1)$.
In implementation, $t=1$ is not sampled because the objective has a singular weight at $t=1$, whereas generation does not require evaluating the drift at the terminal time.
This truncation avoids direct evaluation of quantities that become singular or degenerate at $t=1$.

\subsection{Johnson's $S_U$ prior}
Next, we choose the prior as Johnson's $S_U$ distribution~\cite{johnson1949systems}:
\begin{equation}
  \pi(y) := \frac{1}{\sqrt{2\pi(1+y^2)}}\exp \left[ -\frac{1}{2}(\sinh^{-1} y)^2\right].
  \label{eq:JS_dist}
\end{equation}
This prior is heavy-tailed, is not sub-Gaussian, and differs substantially
from the standard Gaussian.
Sampling from $\pi$ is straightforward because it is the push-forward of the standard Gaussian under the componentwise map $\sinh$:
\begin{equation}
  Y = \sinh \xi, \quad \xi \sim \mathcal{N}(0, I_d).
  \label{eq:sample_JS}
\end{equation}
We use the construction in Sec.~\ref{subsec:general_priors} with a linear interpolation:
\begin{equation}
  \phi_t^x(y) = (1-t)y + tx,
  \label{eq:phi_t^x}
\end{equation}
and $\gamma(t)=(1-t)^{-1}$.
Indeed, this $\phi_t^x$ satisfies the condition in Eq.~\eqref{eq:x-independent_cond} (see also Appendix~\ref{app:x_independent_diffusion}).
Then $b(t,z)=\sqrt{1-t}\,I_d$, as in the Gaussian-prior experiment.
Using the reparameterization trick $Z=tx+(1-t)\sinh\xi$, we obtain the following objective; see Appendix~\ref{app:objective_JS} for the derivation:
\begin{widetext}
  \begin{equation}
    \mathcal{L}_\text{SF}(\theta)
    = \frac{1}{2n}\sum_{i=1}^n\mathbb{E}_{t\sim \text{U}_{[0, 1)}, \xi \sim \mathcal{N}(0, I_d)}\!\left[ \frac{1}{1-t}\left\| x_i - \sinh \xi - \frac{1}{2}\frac{\xi+\tanh \xi}{\cosh \xi} -s_\theta(t, tx_i + (1-t)\sinh \xi)  \right\|^2_{I_d} \right],
    \label{eq:objective_JS}
  \end{equation}
\end{widetext}
where the hyperbolic functions are applied componentwise.

\subsection{Results}
For comparison, we also trained a standard variance-preserving SBM (VP-SBM)~\cite{song2020score} based on the OU reference process:
\begin{equation}
  dX_t = \frac{1}{2}X_t\,dt + dW_t,\quad X_0 \sim \mathcal{N}(0, I_d).
\end{equation}
We used the same network architecture, optimizer, batch size, and number of training iterations as in our method.
For the VP-SBM, the time horizon was set to $T=10$, and both training and generation were performed on this interval.
Generated samples were obtained by the Euler--Maruyama method with $100$ uniform time steps.

\begin{figure}[htbp]
  \centering
  \includegraphics[width=0.8\linewidth]{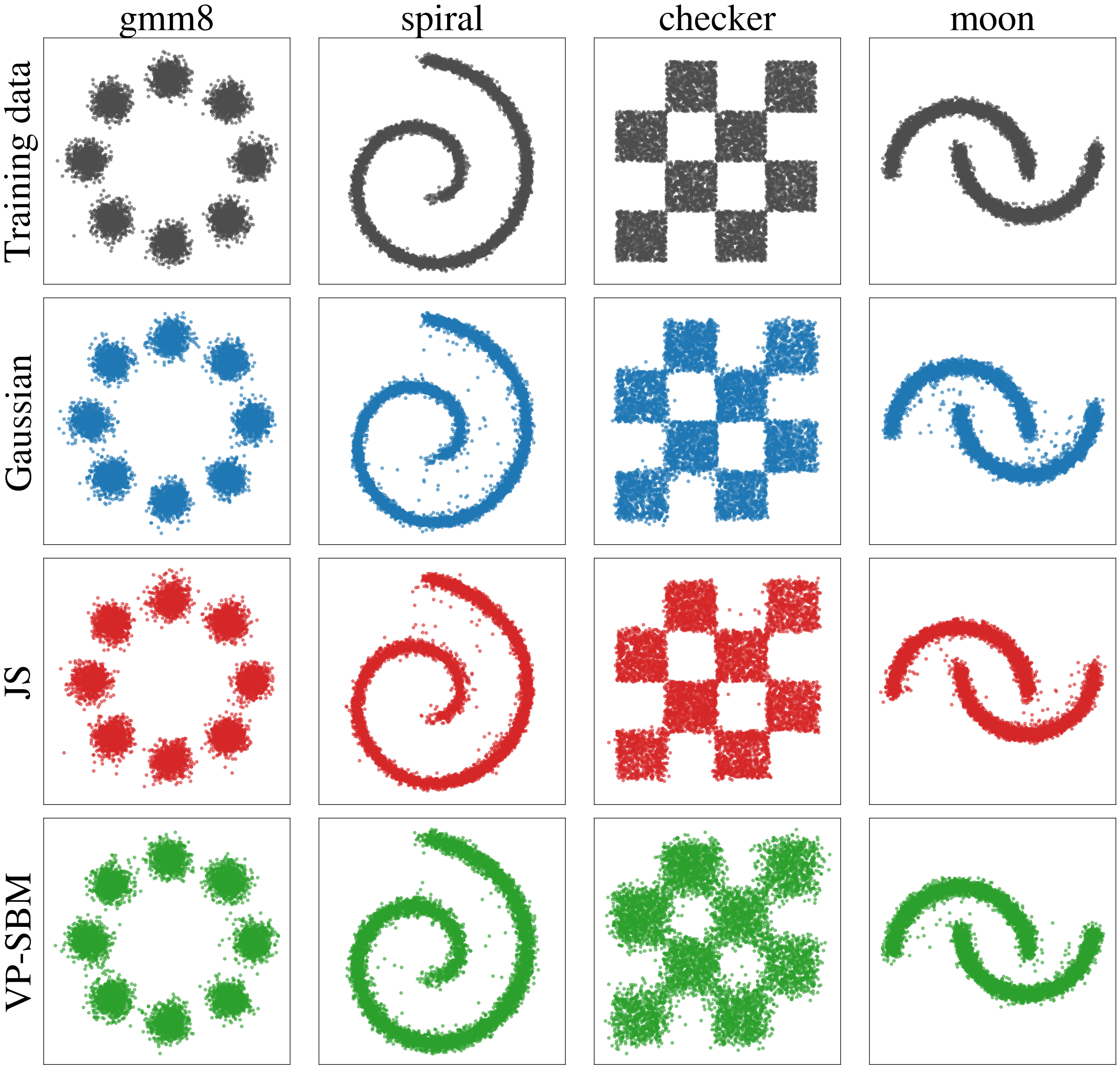}
  \caption{
    The visualization of 7,500 points in training datasets and 7,500 generated samples from trained models.
    The top row represents the training data.
    The second and third rows show samples from the proposed method with the standard Gaussian prior and the Johnson's $S_U$ prior, respectively.
    The bottom row shows samples from the VP-SBM with $T=10$.
    All generated samples were obtained using $100$ Euler--Maruyama steps.
    }
  \label{fig:exp_results}
\end{figure}

Figure~\ref{fig:exp_results} shows the training data and generated samples on four two-dimensional toy datasets.
The proposed method reproduces the qualitative structure of all datasets both with the standard Gaussian prior and with the Johnson's $S_U$ prior.

The VP-SBM with $T=10$ also captures the overall structure of the datasets, but it requires choosing the time horizon $T$.
As shown in Appendix~\ref{app:vp-sbm}, inappropriate choices of $T$ degraded sample quality in our setup.
This reflects the SBM trade-off between convergence to the prior and the discretization accuracy of the Euler--Maruyama method.
The advantage of our framework is that, without this time-horizon tuning, we can well extract the qualitative structure from given datasets.

\section{Conclusion}\label{sec:conclusion}

We introduced a construction principle for the reference process that combines simulation-free training with finite-time generation.
Our construction first prescribes tractable time-dependent conditional distributions and then derives the corresponding SDE. 
This formulation yields an objective based on a path-space KL divergence that can be evaluated without simulating the reference SDE. 
In addition, defining the conditional distributions over the finite interval $[0,1]$ ensures that the resulting process connects the prior and empirical distributions within finite time.

The framework also provided a structural interpretation of existing methods.
It identifies time reversal of a data-to-prior process as the origin of score matching and recovers conditional flow matching as the small-noise limit of our stochastic formulation.
Numerical experiments demonstrated that the proposed construction with both Gaussian and non-Gaussian priors successfully learn the dataset.
They further illustrated that time-horizon tuning required by SBMs is not needed in our construction.

Similar to our construction, Stochastic interpolants (SIs)~\cite{albergo2025stochastic} are known to be the model achieving both simulation-free training and finite-time generation simultaneously.
SIs yield finite-time probability paths from an explicitly prescribed random interpolation. 
For example, let $X_0\sim\pi$, $X_1\sim\mu$, and $\xi\sim\mathcal{N}(0,I_d)$ be independent random variables, and define
\begin{equation}
Z_t=I(t,X_0,X_1)+\gamma(t)\xi,
\end{equation}
where $I(0,X_0,X_1)=X_0$, $I(1,X_0,X_1)=X_1$, and $\gamma(0)=\gamma(1)=0$. 
The time-dependent distribution of $Z_t$ connects $\pi$ and $\mu$ over the finite interval $[0,1]$. 
Its velocity and score fields can be evaluated through quadratic objectives using direct samples of $(X_0,X_1,\xi)$. 
Moreover, the resulting probability path can be realized by both a deterministic flow and a family of diffusion processes with tunable diffusion coefficients. 
Clarifying the precise relation between SIs and our construction beyond these shared properties remains an important direction for future work.

Also, SDE matching~\cite{bartosh2025sde} is recently developed as the generative model for time series data.
Since this model is based on the idea of the flow-based diffusion models reviewed in Sec.~\ref{subsec:flow-based}, it is expected that we can reinterpret this model in our framework.

Another direction for future work is the systematic design of the reference process. 
As demonstrated for several interpolation schedules for Gaussian priors in Appendix~\ref{app:exp_varphi}, the design of $\{\rho_t\}_{t\in[0, 1]}$ affects generation quality even when the boundary distributions are unchanged.
Accordingly, establishing principles for selecting these functional form is necessary to improve training and generation performance.

\begin{acknowledgments}
    This work was supported by 
    JST SPRING Grant No. JPMJSP2106 
    and 
    the Cross-ministerial Strategic Innovation Promotion Program (SIP) from the Cabinet Office (No. 23836436).
\end{acknowledgments}

\appendix

\section{Derivation of the simulation-free objective}
\label{app:objective-SF}
Using Eq.~\eqref{eq:p-decomp}, Eq.~\eqref{eq:objective-2} can be written as
\begin{align}
  \mathcal{L}_\text{DM}(\theta)
  &= \frac{1}{2} \int_0^1 dt \int dz\, p_t(z) \|a(t,z)-s_\theta(t,z) \|^2_{\Gamma^{-1}(t, z)}\nonumber\\
  &= \frac{1}{2} \int_0^1 dt \int dx \int dz\, \mu(x)\rho_t(z|x)\nonumber\\
  &\hspace{1.5cm}\times \|a(t,z)-s_\theta(t,z)\|^2_{\Gamma^{-1}(t, z)}.
  \label{eq:app_path_objective}
\end{align}
The integrand of Eq.~\eqref{eq:app_path_objective} is expanded as 
\begin{equation}
  \begin{aligned}
    &\|a-s_\theta\|^2_{\Gamma^{-1}}\\
    &= \|\alpha-s_\theta\|^2_{\Gamma^{-1}} - \|a-\alpha\|^2_{\Gamma^{-1}} + 2 (a-\alpha)^\top\Gamma^{-1}(a-s_\theta).
    \label{eq:app_norm_identity}
  \end{aligned}
\end{equation}
Since $\alpha$ is defined by Eq.~\eqref{eq:drift_SF}, the last term in Eq.~\eqref{eq:app_norm_identity} vanishes after integration with respect to $x$:
\begin{align}
    &\int dx\, \mu(x)\rho_t(z|x)\nonumber\\
    &\quad\times(a(t,z)-\alpha(t,z,x))^\top\Gamma^{-1}(t,z)(a(t, z) - s_\theta(t, z))\nonumber\\
    &= \left[p_t(z)a(t,z) - \int dx\, \mu(x)\rho_t(z|x)\alpha(t,z,x)\right]^\top \Gamma^{-1}(t, z)\nonumber\\
    &\quad \times (a(t, z) - s_\theta(t, z))\nonumber\\
    &=0.
    \label{eq:app_zero_term}
\end{align}
Therefore, the substitution of Eq.~\eqref{eq:app_norm_identity} into Eq.~\eqref{eq:app_path_objective} can be calculated as
\begin{align}
  \mathcal{L}_\text{DM}(\theta) 
  = &\frac{1}{2} \int_0^1 dt \int dx \int dz\, \mu(x)\rho_t(z|x)\nonumber\\
  &\qquad\quad\times \|\alpha(t,z,x)-s_\theta(t,z)\|^2_{\Gamma^{-1}(t, z)}. \nonumber\\ 
  &- \frac{1}{2} \int_0^1 dt \int dx \int dz\, \mu(x)\rho_t(z|x)\nonumber\\
  &\qquad\quad\times\|a(t,z)-\alpha(t,z,x)\|^2_{\Gamma^{-1}(t, z)}.
  \label{eq:app_objective_relation}
\end{align}
Since the second term in Eq.~\eqref{eq:app_objective_relation} is independent of $\theta$, we define the first term as an effective objective:
\begin{equation}
  \begin{aligned}  
    &\mathcal{L}_\text{SF}(\theta):=\\
    &\frac{1}{2}\int_0^1 dt \,\mathbb{E}_{Z \sim \rho_t(\cdot|X), X\sim \mu}\!\Big[\|\alpha(t, Z, X) - s_\theta(t, Z)\|^2_{\Gamma(t, Z)^{-1}}\Big].
  \end{aligned}
\end{equation}
Note that the following identity obviously holds
\begin{align}
    \arg\min_\theta \mathcal{L}_\text{DM}(\theta)
    = \arg\min_\theta \mathcal{L}_{\mathrm{SF}}(\theta).
\end{align}
Thus, minimizing the path-space objective in Eq.~\eqref{eq:objective-2} is equivalent to minimizing the simulation-free objective in Eq.~\eqref{eq:objective_SF}.

\section{Derivation of the reference SDE for the Gaussian priors}
\label{app:gaussian-coefficients}
We derive Eqs.~\eqref{eq:alpha_gauss} and \eqref{eq:b_gauss} under the assumption that $b$ is independent of $z$ and isotropic: $b(t,z) = c(t)I_d$ where $c:[0,1]\to\mathbb{R}_{>0}$ is arbitrary. 
Accordingly, the diffusion covariance is
\begin{equation}
\Gamma(t)=c(t)^2I_d.
\end{equation}

The conditional distribution $\rho_t$ in Eq.~\eqref{eq:conds_gauss} is represented as
\begin{equation}
  \rho_t(z|x) = \frac{1}{(2\pi\sigma(t)^2)^{d/2}} \exp\left[ -\frac{1}{2}\frac{\|z-m_t(x)\|^2_{I_d}}{\sigma(t)^2} \right].
\end{equation}
To evaluate the conditional FPE~\eqref{eq:cond-FPE}, we first compute the derivatives
of $\rho_t$:
\begin{align}
  \frac{\partial_t\rho_t(z|x)}{\rho_t(z|x)} 
  &= - d\frac{\dot{\sigma}(t)}{\sigma(t)} 
  + \frac{(z-m_t(x))\cdot \partial_tm_t(x)}{\sigma(t)^2}\nonumber\\ 
  &\hfill+ \frac{\dot{\sigma}(t)}{\sigma(t)^3}\|z-m_t(x)\|^2_{I_d},
  \label{eq:app-del_t_p}\\
  \frac{\nabla_z\rho_t(z|x)}{\rho_t(z|x)} 
  &= -\frac{z-m_t(x)}{\sigma(t)^2},
  \label{eq:app-del_z_p}\\
  \frac{\nabla_z^2\rho_t(z|x)}{\rho_t(z|x)}
  &= \frac{\|z-m_t(x)\|^2_{I_d}}{\sigma(t)^4} - \frac{d}{\sigma(t)^2}.
  \label{eq:app-del_z^2_p}
\end{align}
Since $\Gamma(t)=c(t)^2I_d$ is independent of $z$, dividing the FPE~\eqref{eq:cond-FPE} by $\rho_t(z|x)$ gives
\begin{align}
  \frac{\partial_t\rho_t}{\rho_t}
  = -\nabla_z\cdot\alpha - \alpha\cdot \frac{\nabla_z\rho_t}{\rho_t} + \frac{c(t)^2}{2}\frac{\nabla_z^2\rho_t}{\rho_t}.
  \label{eq:app_condFPE_divided}
\end{align}
Furthermore, the substitution of Eqs.~\eqref{eq:app-del_t_p}--\eqref{eq:app-del_z^2_p} into Eq.~\eqref{eq:app_condFPE_divided}, we obtain the partial differential equation (PDE) for $\alpha$:
\begin{align}
  &-\nabla_z\cdot\alpha(t,z,x) + \frac{z-m_t(x)}{\sigma(t)^2}\cdot \alpha(t,z,x)\nonumber\\
  &= - d\frac{\dot{\sigma}(t)}{\sigma(t)} + \frac{(z-m_t(x))\cdot \partial_tm_t(x)} {\sigma(t)^2} + \frac{\dot{\sigma}(t)} {\sigma(t)^3}\|z-m_t(x)\|^2 \nonumber\\ 
  &\qquad - \frac{1}{2}\frac{c(t)^2}{\sigma(t)^4} \|z-m_t(x)\|^2_{I_d} + \frac{d}{2}\frac{c(t)^2}{\sigma(t)^2}.
  \label{eq:app-alpha-equation}
\end{align}
By solving this PDE~\eqref{eq:app-alpha-equation} under the boundary condition
\begin{equation}
  \lim_{|z|\to\infty} \alpha(t,z,x)\rho_t(z|x)=0,
\end{equation}
we get
\begin{align}
  \alpha(t,z,x) = \partial_tm_t(x) + \left[ \frac{\dot{\sigma}(t)}{\sigma(t)} - \frac{1}{2}\frac{c(t)^2}{\sigma(t)^2} \right](z-m_t(x)).
  \label{eq:app-alpha-b}
\end{align}
Here, we note that $c(t)$ is an arbitrary function.
By scaling $c(t)$ by $\sigma(t)$ and defining $\lambda(t) := c(t)^2/\sigma(t)^2$, we get
\begin{equation}
  \alpha(t,z,x) 
  = \partial_tm_t(x) + \left[ \frac{\dot{\sigma}(t)}{\sigma(t)} - \frac{1}{2}\lambda(t) \right] (z-m_t(x)),
\end{equation}
and
\begin{equation}
  b(t) = \sqrt{\lambda(t)}\,\sigma(t)I_d.
\end{equation}
These are Eqs.~\eqref{eq:alpha_gauss} and \eqref{eq:b_gauss}.

\section{Condition for an $x$-independent diffusion coefficient}
\label{app:x_independent_diffusion}
The diffusion coefficient in Eq.~\eqref{eq:general_b} is
\begin{equation}
  \beta(t,z,x) = \sqrt{\gamma(t)} D_y\phi_t^x((\phi_t^x)^{-1}(z)),
\end{equation}
where $D_y\phi_t^x$ denotes the Jacobian matrix.

We required $\beta$ to be independent of $x$: $\nabla_x \beta(t,z,x)=0$.
By the inverse function theorem,
\begin{equation} 
  D_y\phi_t^x\left((\phi_t^x)^{-1}(z)\right) = [D_z(\phi_t^x)^{-1}(z)]^{-1}.
\end{equation}
Hence, the condition above implies that $D_z(\phi_t^x)^{-1}(z)$ is independent of $x$.
Fix an arbitrary $x_0\in\mathbb{R}^d$ and define a smooth bijection $f_t(y) := \phi_t^{x_0}(y)$.
Since the Jacobians of $(\phi_t^x)^{-1}$ and $f_t^{-1}$ with respect to $z$ are identical, their difference is independent of $z$.
Thus, there exists a function $g_t:\mathbb{R}^d\to\mathbb{R}^d$ such that 
\begin{equation}
  (\phi_t^x)^{-1}(z) = f_t^{-1}(z)-g_t(x).
\end{equation}
Setting $y=(\phi_t^x)^{-1}(z)$ gives
\begin{equation} 
  \phi_t^x(y) = f_t(y+g_t(x)).
\end{equation}

Conversely, suppose that
\begin{equation}
  \phi_t^x(y) = f_t(y+g_t(x)),
\end{equation}
where $f_t:\mathbb{R}^d\to\mathbb{R}^d$ is a smooth bijection. 
By setting $z = \phi_t^x(y)$, we have
\begin{equation}
  y + g_t(x) = f_t^{-1}(z).
\end{equation}
Therefore,
\begin{equation}  
  D_y\phi_t^x(y) = D_y f_t(y+g_t(x)) = D_y f_t(f_t^{-1}(z)),
\end{equation}
which is independent of $x$. Consequently,
\begin{equation}
  b(t,z) = \beta(t, z, x) = \sqrt{\gamma(t)} D_y f_t(f_t^{-1}(z)).
\end{equation}

Thus, the family
\begin{equation}
  \phi_t^x(y)=f_t(y+g_t(x))
\end{equation}
yields an $x$-independent diffusion coefficient.

The linear interpolation in Eq.~\eqref{eq:phi_t^x} is included in this class.
It is obtained by choosing
\begin{equation} 
  f_t(y)=(1-t)y, \quad g_t(x)=\frac{t}{1-t}x.
\end{equation}
Indeed,
\begin{equation}
  \phi_t^x(y) = f_t(y+g_t(x)) = (1-t)y+tx.
\end{equation}

\section{Derivation of the practical objective functions}
We derive two objective functions used in Sec.~\ref{sec:numerical} as follows.

\subsection{Gaussian prior}\label{app:objective_gauss}
By substituting Eq.~\eqref{eq:phi_prac} and $\lambda(t) = \sigma(t)^{-1}$ into Eqs.~\eqref{eq:alpha_gauss} and \eqref{eq:b_gauss}, we obtain
\begin{gather}
  \begin{aligned}
    \alpha(t, z, x) 
    &= x + \left( -\frac{1}{1-t} - \frac{1}{2}\frac{1}{1-t} \right)(z-tx)\\
    &= x -\frac{3}{2}\frac{z-tx}{1-t},
  \end{aligned}\\
  b(t, z) = \sqrt{1-t}\,I_d.
\end{gather}
By employing the reparameterization trick in Eq.~\eqref{eq:rep_trick} into the above equation, we get $\alpha$ as a function of $\xi$ instead of $z$:
\begin{align}
  \alpha(t, tx + (1-t)\xi, x) = x - \frac{3}{2}\xi.
\end{align}
Therefore, the expectation value in Eq.~\eqref{eq:objective_SF} can be represented as
\begin{equation}
  \begin{aligned}
    &\mathbb{E}_{Z \sim \rho_t(\cdot|x_i)}\!\left[\|\alpha(t, Z, x_i) - s_\theta(t, Z)\|^2_{\Gamma^{-1}(t,Z)}\right]\\
    &= \mathbb{E}_{\xi \sim \mathcal{N}(0, I_d)}\!\Bigg[\frac{1}{1-t}\bigg\| x - \frac{3}{2}\xi - s_\theta(t, tx + (1-t)\xi)\bigg\|^2_{I_d}\Bigg].
  \end{aligned}
\end{equation}
Moreover, we approximately compute the time integral in Eq.~\eqref{eq:objective_SF} by Monte Carlo integration with a uniform distribution over $[0, 1)$.
As a result, we get the objective function Eq.~\eqref{eq:objective_gaussian}.

\subsection{Johnson's $S_U$ prior}\label{app:objective_JS}
From the definition of the Johnson's $S_U$ distribution Eq.~\eqref{eq:JS_dist}, the gradient of its $\log$-density is
\begin{align}
  \nabla_y \log \pi(y) = - \frac{\sinh^{-1} y}{\sqrt{1+y^2}} - \frac{y}{1 + y^2}.
\end{align}
Substituting Eq.~\eqref{eq:phi_t^x} and $\gamma(t) = 1/(1-t)$ into Eqs.~\eqref{eq:general_alpha} and \eqref{eq:general_b} gives
\begin{align}
  \alpha(t, tx + (1-t)y, x) = x - y - \frac{1}{2}\left( \frac{\sinh^{-1} y}{\sqrt{1+y^2}} + \frac{y}{1 + y^2} \right).
\end{align}
Also, $b(t, z) = \sqrt{1-t}I_d$. 
Furthermore, the reparameterization trick in Eq.~\eqref{eq:sample_JS} gives
\begin{align}
  &\alpha(t, tx + (1-t)\sinh \xi, x)\nonumber\\ 
  &= x - \sinh \xi - \frac{1}{2}\left( \frac{\xi}{\sqrt{1+\sinh^2 \xi}} + \frac{\sinh \xi}{1 + \sinh^2 \xi} \right)\nonumber\\
  &= x - \sinh \xi - \frac{1}{2}\ \frac{\xi + \tanh \xi}{\cosh \xi}.
\end{align}
Following the same procedure as in the Gaussian-prior case, we obtain Eq.~\eqref{eq:objective_JS}.

\section{Additional experiments}\label{app:add_exp}

\subsection{Effect of the time horizon in VP-SBM}\label{app:vp-sbm}
\begin{figure}[htbp]
  \centering
  \includegraphics[width=\linewidth]{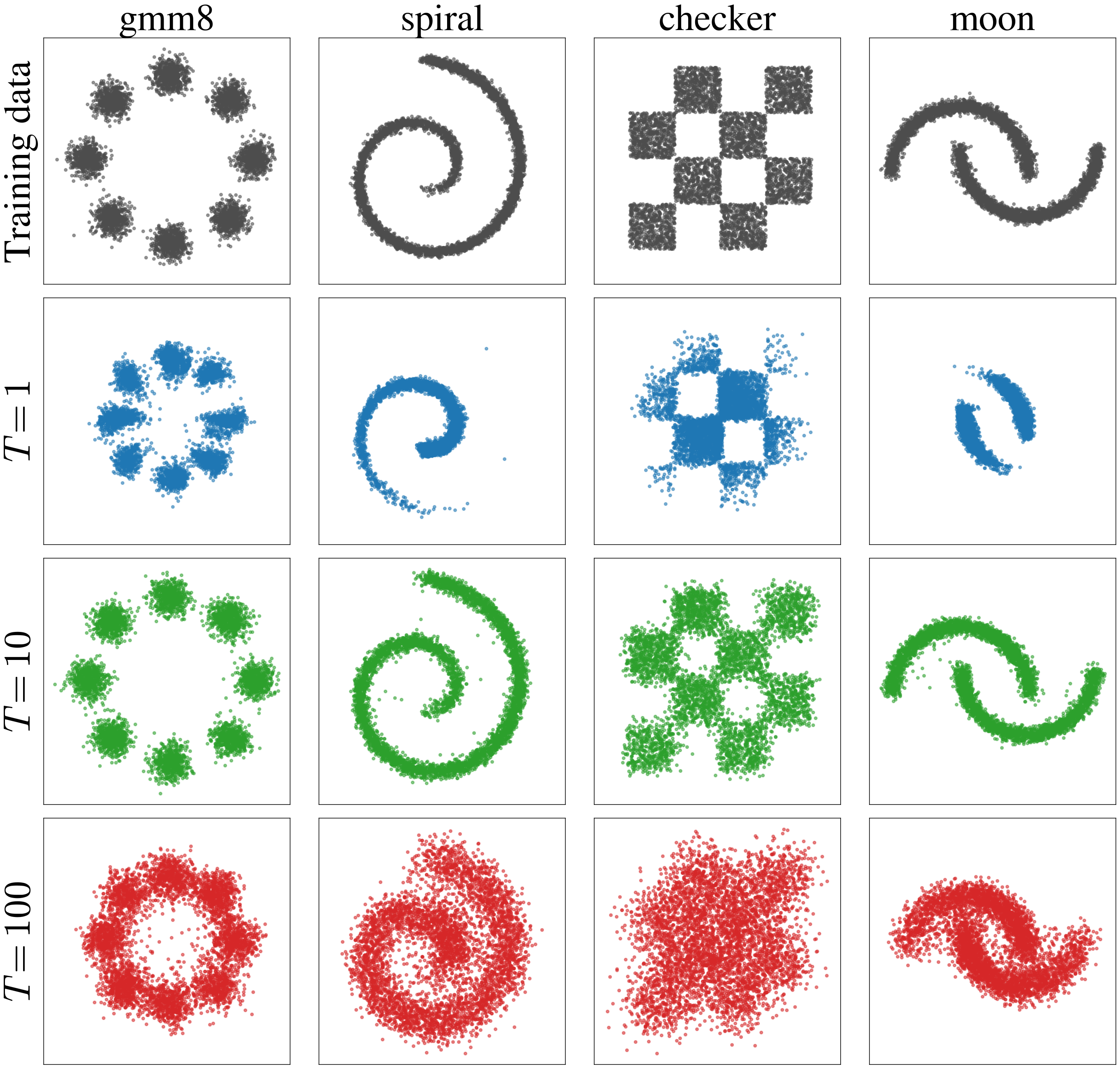}
  \caption{
    The visualization of 7,500 training data and generated samples from VP-SBMs.
    The top row represents the training data.
    The remaining rows show samples from the VP-SBM with $T=1, 10, 100$, respectively.
    All generated samples were obtained using $100$ Euler--Maruyama steps.
    Small $T$ leads to insufficient training, whereas large $T$ degrades the SDE simulation for sampling.
    }
  \label{fig:vp-sbm}
\end{figure}

We examined the sensitivity of the VP-SBM to the time horizon $T$.
Figure~\ref{fig:vp-sbm} shows generated samples for $T=1,10,100$, where the number of discretization steps (i.e. Euler--Maruyama steps) was fixed to $100$ in all cases, as in the setting in Sec.~\ref{sec:numerical}.

The results for different $T$ reveal a trade-off between training and generation in SBMs.
For $T=1$, the first term in Eq.~\eqref{eq:ESM} remains nonzero because the initial distribution of the reference process has not yet sufficiently approached the prior. 
Consequently, we never make the KL divergence sufficiently small (see Eqs.~\eqref{eq:objective} and \eqref{eq:ESM}).
For $T=100$, while the first term in Eq.~\eqref{eq:ESM} is negligible, discretization error is increasing. 
For a fixed number of discretization steps $N$, the size of each time step is proportional to $T/N$ and therefore increases with $T$. 
Therefore, increasing $T$ under the fixed discretization steps reduces the accuracy of the SDE simulation, which causes the degradation of the quality of the generated samples.
In contrast, for $T=10$, this trade-off is mitigated. 

These results show that, in SBMs with a fixed number of discretization steps, the time horizon induces a trade-off between the training efficiency achieved through convergence to the prior and the numerical accuracy of generation with the SDE simulation. 
It is therefore necessary to tune the time horizon to balance these two sources of error.

\subsection{Effect of the interpolation schedule}\label{app:exp_varphi}

We demonstrate that the performance depends on the design of the reference process realizing the prescribed conditional distributions, even when those distributions are identical.

\begin{figure}[htbp]
    \centering
    \includegraphics[width=0.8\linewidth]{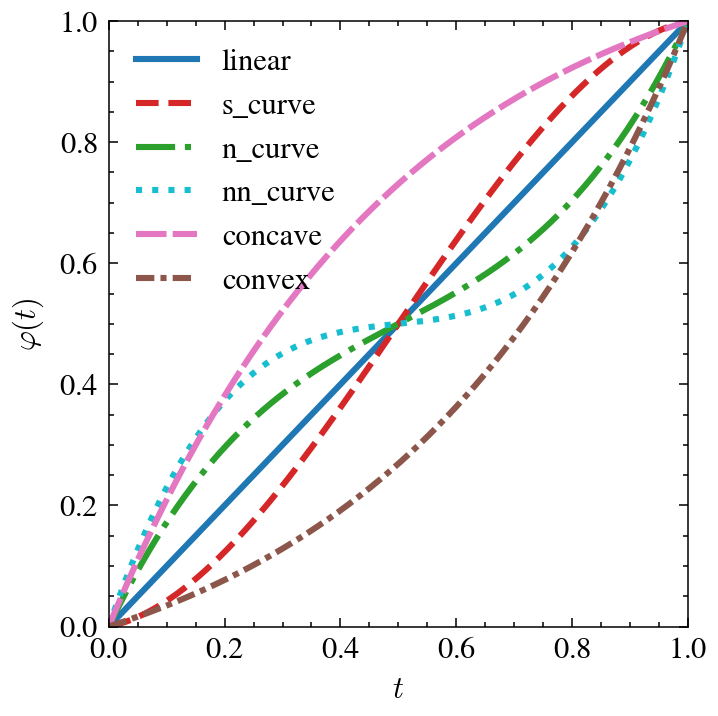}
    \caption{Interpolation schedules $\varphi(t)$ used in the additional experiments. All schedules satisfy $\varphi(0)=0$ and $\varphi(1)=1$, but differ in their intermediate time dependence.}
    \label{fig:varphi}
  \end{figure}

The construction for Gaussian priors in Sec.~\ref{subsec:gaussian_prior} contains freedom in the time scheduling of the mean $m_t$ and the standard deviation $\sigma(t)$.
To examine the effect of this schedule, we parameterize both functions by a monotone function $\varphi: [0,1]\to [0,1]$:
\begin{equation}
  m_t(x) = \varphi(t)x, \quad \sigma(t) = 1-\varphi(t).
  \label{eq:schedule}
\end{equation}
Here, to satisfy Eq.~\eqref{eq:bdry_gauss}, $\varphi$ satisfies $\varphi(0)=0$ and $\varphi(1)=1$.
We consider the linear schedule $\varphi(t) = t$ in Eq.~\eqref{eq:phi_prac} together with cubic and exponential families:
\begin{equation}
    \varphi(t) = (1-c)t+3ct^2-2ct^3, \quad \varphi(t) = \frac{e^{ct}-1}{e^c-1},
\end{equation}
where $c$ is an arbitrary parameter.
The values of the parameter $c$ are summarized in Table~\ref{tab:varphi}, and the corresponding schedule curves are shown in Fig.~\ref{fig:varphi}.
\begin{table}[htbp]
    \centering
    \caption{Time schedules $\varphi(t)$ used in the Gaussian-prior experiments. The table lists the functional form and parameter $c$ for each schedule. All schedules satisfy $\varphi(0)=0$ and $\varphi(1)=1$.}
    \label{tab:varphi}
    \begin{tabular}{lcc}
        \hline
        Name & $\varphi(t)$ & $c$ \\
        \hline
        \hline
        \texttt{linear} & $t$ & \\
        \hline
        \texttt{s\_curve} & \multirow{3}{*}{$(1-c)t+3ct^2-2ct^3$} & $0.8$ \\ 
        \texttt{n\_curve} &  & $-1$ \\ 
        \texttt{nn\_curve} &  & $-1.8$ \\
        \hline
        \texttt{concave} & \multirow{2}{*}{$\dfrac{e^{ct}-1}{e^c-1}$} & $-2$ \\ 
        \texttt{convex} &  & $2$ \\
        \hline
    \end{tabular}
\end{table}

Substituting Eq.~\eqref{eq:schedule} and $\lambda(t) = \sigma(t)^{-1}$ into Eqs.~\eqref{eq:alpha_gauss} and \eqref{eq:b_gauss}, and using the reparameterization trick $Z=\varphi(t)x+(1-\varphi(t))\xi$, the simulation-free objective in Eq.~\eqref{eq:objective_SF} can be calculated as
\begin{widetext}
  \begin{equation}
    \mathcal{L}_\text{SF}(\theta) 
    = \frac{1}{2n}\sum_{i=1}^n\mathbb{E}_{t\sim \text{U}_{[0, 1)}, \xi \sim \mathcal{N}(0, I_d)}\!\left[ 
      \frac{1}{1-\varphi(t)} \left\| 
        \dot{\varphi}(t)x_i - \left\{\dot{\varphi}(t) + \frac{1}{2} \right\}\xi - s_\theta(t, \varphi(t)x_i + (1-\varphi(t))\xi)
        \right\|^2_{I_d}
        \right].
  \end{equation}
\end{widetext}
Although all schedules connect the same boundary distributions $\mu$ and $\pi$, they therefore define different training objectives and lead to different generation dynamics.
As in the main experiments, the time variable was sampled uniformly from $[0,0.99]$ in implementation.

\begin{table}[htbp]
\centering
\caption{The mean of MMD over 10 independent training runs with
different random seeds for each dataset and curve. Values are reported in units of $10^{-4}$. The best value is bold and underlined, and the second-best value is bold.}
\label{tab:mmd_highlighted}
\setlength{\tabcolsep}{5pt}
\begin{tabular}{lcccc}
\hline
Curve & gmm8 & spiral & checker & moon \\
\hline
\hline
\texttt{linear} 
& \underline{\textbf{0.989}}
& \underline{\textbf{5.46}}
& \textbf{6.11}
& \underline{\textbf{6.05}} \\
\texttt{s\_curve} 
& 3.65 & 9.31 & 9.62 & 11.0 \\
\texttt{n\_curve} 
& \textbf{1.63} & \textbf{6.28} & 6.56 & 12.1 \\
\texttt{nn\_curve} 
& 3.47 & 7.00 & 7.53 & 16.1 \\
\texttt{concave} 
& 1.59 & 5.57 & \underline{\textbf{4.39}} & 12.6 \\
\texttt{convex} 
& 2.19 & 9.07 & 8.42 & \textbf{10.3} \\
\hline
\end{tabular}
\end{table}

\begin{table}[htbp]
\centering
\caption{The mean of SWD over 10 independent training runs with
different random seeds for each dataset and curve.  Values are reported in units of $10^{-1}$. The best value is bold and underlined, and the second-best value is bold.}
\label{tab:swd_highlighted}
\setlength{\tabcolsep}{5pt}
\begin{tabular}{lcccc}
\hline
Curve & gmm8 & spiral & checker & moon \\
\hline
\hline
\texttt{linear} 
& \underline{\textbf{3.15}} 
& \underline{\textbf{3.04}} 
& \textbf{1.97} 
& \underline{\textbf{4.09}} \\

\texttt{s\_curve} 
& 4.46 
& 3.83 
& 2.53 
& 5.2 \\

\texttt{n\_curve} 
& \textbf{3.3} 
& 3.38 
& 2.06 
& 5.27 \\

\texttt{nn\_curve} 
& 4.27 
& \textbf{3.27} 
& 2.11 
& 6.08 \\

\texttt{concave} 
& 3.5 
& 3.39 
& \underline{\textbf{1.76}} 
& 5.23 \\

\texttt{convex} 
& 3.49 
& 3.57 
& 2.29 
& \textbf{5.14} \\
\hline
\end{tabular}
\end{table}
We trained the model with each interpolation schedule on the four datasets used in Sec.~\ref{sec:numerical}.
Tables~\ref{tab:mmd_highlighted} and~\ref{tab:swd_highlighted} report an average of the maximum mean discrepancy (MMD)~\cite{gretton2012kernel} and sliced Wasserstein distance (SWD)~\cite{bonneel2015sliced} over 10 independent training runs with different random seeds, respectively.
Lower values indicate better agreement between the generated and training distributions.

The linear schedule achieved the lowest MMD and SWD on the Gaussian-mixture, spiral, and two-moons datasets.
For the checkerboard dataset, the concave exponential schedule performed best under both metrics.
These results show that the intermediate conditional path can affect generation quality even when the boundary distributions are unchanged.
The linear schedule provides a robust choice in the present experiments, whereas the checkerboard results suggest that a suitably chosen nonlinear schedule can be advantageous for some target distributions.
A systematic characterization or learning of the interpolation schedule therefore constitutes an important direction for future work.

\bibliography{99_references.bib}

\end{document}